\documentclass[runningheads]{llncs}
\usepackage{graphicx}
\usepackage{cite}
\usepackage{amsmath,amssymb,amsfonts}
\usepackage{algorithmic}
\usepackage{textcomp}
\usepackage{hyperref}
\usepackage{fontawesome}
\usepackage{multirow}
\usepackage[utf8]{inputenc}
\usepackage{algorithm}
\usepackage{array}
\usepackage[table]{xcolor}
\usepackage{booktabs}
\usepackage[utf8]{inputenc}
\usepackage{amssymb}

\begin{document}
\title{Editing Knowledge Representation of Language Model via Rephrased Prefix Prompts}
\author{Yuchen Cai\inst{1,2} \and
Ding Cao\inst{1,2} \and
Rongxi Guo\inst{1,2}\and
Yaqin Wen\inst{1,2}\and \\
Guiquan Liu\inst{1,2\textsuperscript{(\faEnvelopeO)}}\and
Enhong Chen\inst{1,2}}
\institute{University of Science and Technology of China, Hefei, China \and
State Key Laboratory of Cognitive Intelligence, Hefei, China\\
\email{\{caiyuchen, caoding, guorongxi, wyq\_65\}@mail.ustc.edu.cn, \{gqliu,cheneh\}@ustc.edu.cn}}

\maketitle              % typeset the header of the contribution
\begin{abstract}
Neural language models (LMs) have been extensively trained on vast corpora to store factual knowledge about various aspects of the world described in texts. Current technologies typically employ knowledge editing methods or specific prompts to modify LM outputs. However, existing knowledge editing methods are costly and inefficient, struggling to produce appropriate text. Additionally, prompt engineering is opaque and requires significant effort to find suitable prompts. To address these issues, we introduce a new method called PSPEM (\textbf{P}refix \textbf{S}oft-\textbf{P}rompt \textbf{E}diting \textbf{M}ethod), that can be used for a lifetime with just one training. It resolves the inefficiencies and generalizability issues in knowledge editing methods and overcomes the opacity of prompt engineering by automatically seeking optimal soft prompts. Specifically, PSPEM adopts a prompt encoder and an encoding converter to compress and refine key information in prompts and adopts prompt alignment techniques to guide model generation, ensuring text consistency and adherence to the intended structure and content. We have validated the effectiveness of PSPEM through knowledge editing and attribute inserting. On the COUNTERFACT dataset, PSPEM achieved nearly 100\% editing accuracy and demonstrated the highest level of fluency. We further analyzed the similarities between PSPEM and original prompts and their impact on the model's internals. The results indicate that PSPEM can serve as an alternative to original prompts, supporting the model in effective editing.

\end{abstract}
\section{Introduction}

\begin{figure}[htb]
\begin{minipage}[b]{1.0\linewidth}
  \centering
  \centerline{\includegraphics[width=12cm]{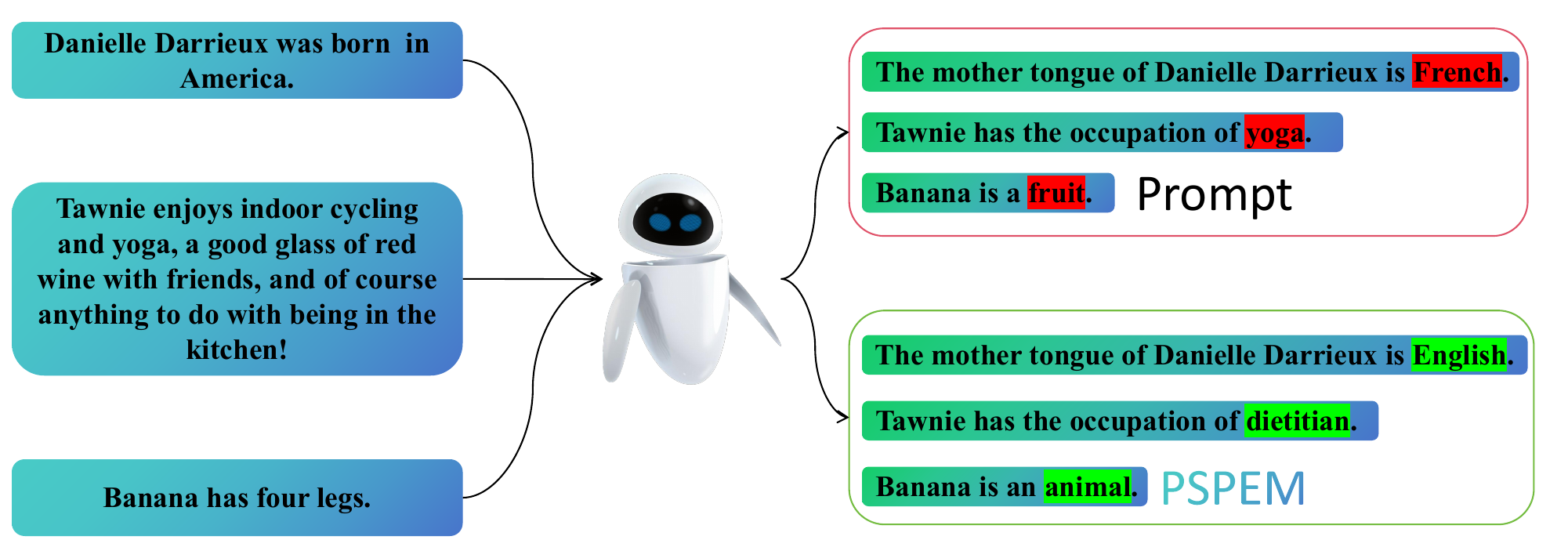}}
%  \vspace{2.0cm}
\end{minipage}
\caption{By inputting the prompts on the left side into the model, traditional prompt engineering generate erroneous text, while PSPEM can correct such errors.}
\label{figure1}
\end{figure}

Language models based on the Transformer architecture, such as GPT and BERT, have revolutionized various natural language processing tasks with their capacity to store and utilize real-world factual knowledge within their parameters \cite{vaswani2017attention,devlin2018bert,radford2019language}. For example, when asked, ``Where is the Eiffel Tower located?" GPT provides the accurate answer, ``Paris". However, inconsistencies or biases present in the pre-training data can propagate into the text generated by the model, leading to errors or contradictions. Additionally, the knowledge of the world changes over time, which presents a challenge to the static knowledge in the model. Addressing this issue requires a nuanced approach to updating the model's knowledge base. Merely retraining the entire model with new data is prohibitively expensive and time-consuming, while fine-tuning, focusing solely on specific updated knowledge poses the risk of overfitting and compromising the model's ability to generalize.

Recent advancements propose a more dynamic and efficient approach to knowledge updating to mitigate these issues, called knowledge editing \cite{wang2023knowledge, yao2023editing}. This technology allows for selective updates and adjustments to the model's knowledge without retraining. These methods aim to balance the need for accurate, up-to-date information with the practical constraints of computational resources and time. The efficacy of knowledge editing is predominantly quantified by two pivotal metrics: generalization and specificity. Generalization entails the model's proficiency in extending the modified knowledge across a spectrum of analogous prompts, ensuring consistent application and understanding of the targeted information \cite{zhu2020modifying,de2021editing}. Conversely, specificity, also referred to as locality, necessitates the model's capacity to isolate the modification impact, safeguarding the unaltered knowledge from inadvertent alteration \cite{mitchell2021fast}. Several new benchmarks are attempting to assess the model's ability to reason with new knowledge, and they are extending these methodologies into the realms of knowledge graphs \cite{cheng2023editing} and multimodal domains \cite{cheng2023can}.

Methods of model editing fall into two distinct classifications depending on the alteration of the original model weights: weight-preserved and weight-modified methods \cite{yao2023editing}. Weight-preserved strategies typically necessitate the inclusion of extra content, while weight-modified techniques directly alter the model's weights. Weight modification methods include hypernetwork-based learning methods and direct optimization methods. Hypernetwork-based learning methods, such as KE \cite{de2021editing}, MEND \cite{mitchell2021fast} and MALMEN \cite{tan2023massive}, utilize a hypernetwork to predict essential updates to the model's weights. Although this technique is promising, it necessitates considerable computational investment for hypernetwork training and frequently diminishes in effectiveness with the increase in language model size \cite{yao2023editing}. Optimization method ROME \cite{meng2022locating} employs causal mediation analysis to identify the editing region and focuses on altering specific information via rank-one adjustments to individual matrices. MEMIT \cite{meng2022mass} adhered to a similar methodology, adeptly modifying several parameter matrices concurrently to facilitate the simultaneous alteration of 10,000 knowledge entities, and showcasing robust generalization and specificity. PMET \cite{li2023pmet} advanced this technique, refining MEMIT's capabilities for more precise editing. However, previous research indicates that minor modifications to the parameters of large language models can impact the model's ultimate behaviour \cite{sun2022black}, and these methods do not allow the model to use new knowledge for reasonable inference \cite{hernandez2023measuring,onoe2023can}.

Prompt engineering \cite{liu2023pre, liu2021gpt} enables the modification of models without necessitating extensive retraining. Altering input prompts, allows models to adapt to diverse tasks and domains, thereby conserving resources. Nonetheless, finding the most effective prompts typically requires considerable manual intervention and iterative experimentation, which can be time-consuming and inefficient. Moreover, discrepancies between the knowledge encapsulated in the prompts and the model's inherent knowledge can lead to erroneous or inconsistent outputs. As depicted in Figure \ref{figure1}, the language model (LM) is prompted with ``Danielle Darrieux was born in America." Upon generating a continuation of this prompt, the LM erroneously asserts that Danielle Darrieux's native language is French, thereby contradicting the prior context. This error occurred because LM developed a memory for Danielle Darrieux's native language, French, during pre-training \cite{mundler2023self,bang2023multitask}.

In-context learning \cite{brown2020language} is a paradigm that does not require retraining, where knowledge is acquired from directly connected demonstrations in the input context. Unlike traditional prompting engineering, the method is capable of learning contextual relationships from multiple given instances, enabling context-based model editing and providing an efficient, lightweight knowledge editing approach \cite{zheng2023can}. Although this method addresses the issue of inconsistent contextual information in the model, it requires searching for multiple guiding instances, which puts an additional burden on knowledge editing. REMEDI \cite{hernandez2023measuring} injects domain-specific knowledge into language models by encoding factual prompts corresponding to knowledge attributes in the direction space. However, when factual prompts conflict with the knowledge already present within the model, the REMEDI method still struggles to handle such contradictions, resulting in inconsistencies or errors in the results.

\begin{figure*}[htb]
\begin{minipage}[b]{\linewidth}
  \centering
  \centerline{\includegraphics[width=12cm]{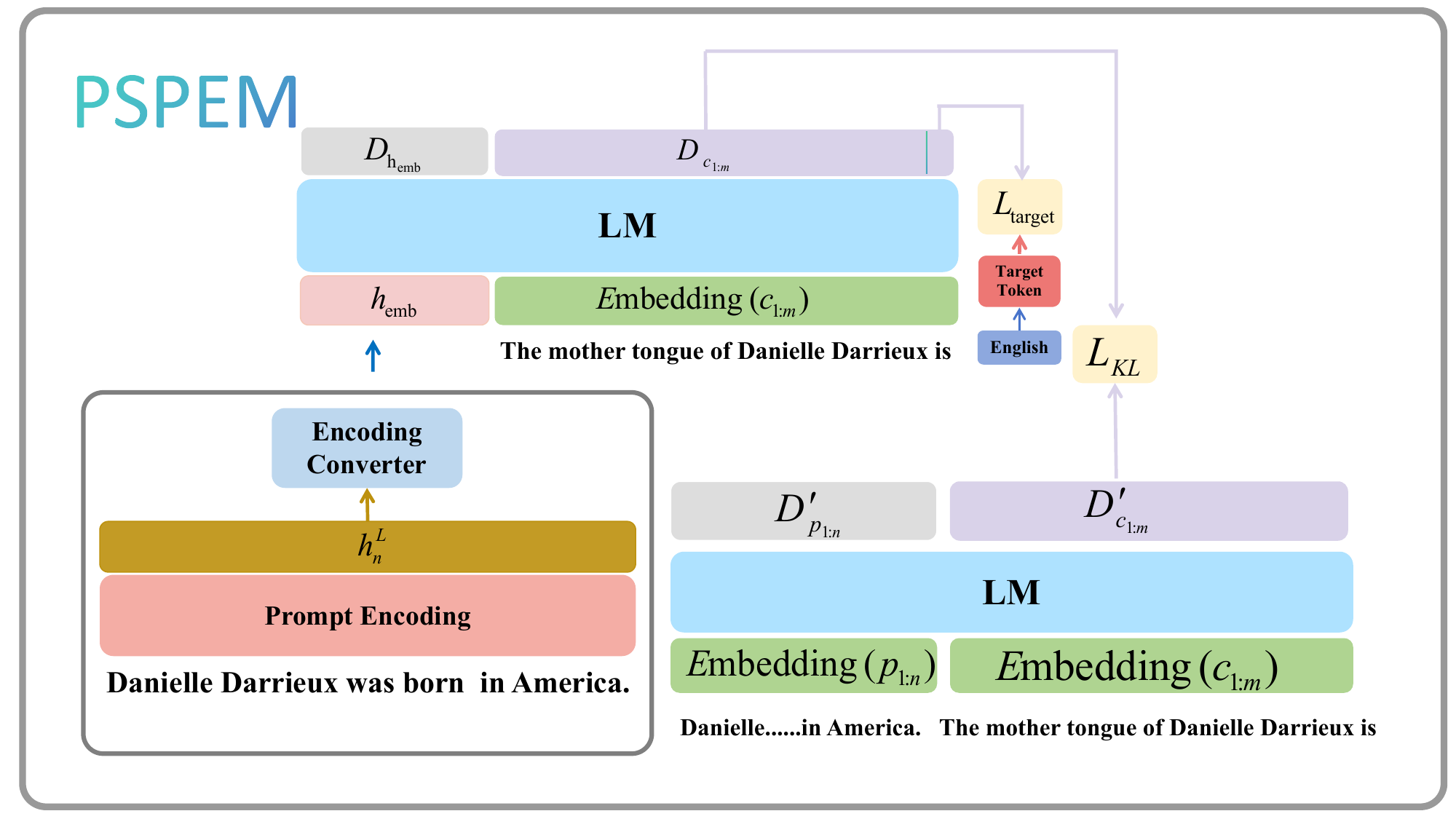}}
\end{minipage}
\caption{Illustration of PSPEM. Given a knowledge prompt (Danielle Darrieux was born in America.) and continuation words (The mother tongue of Danielle Darrieux is), PSPEM constructed more accurately encoded information from the prompt to increase the probability of the target token (English).}
\label{figure2}
\end{figure*}

To overcome the limitation of the weight-modified method in utilizing new knowledge for reasoning and the poor editing accuracy and laborious of the weight-preserved method, we proposed PSPEM (\textbf{P}refix \textbf{S}oft-\textbf{P}rompt \textbf{E}diting \textbf{M}ethod), an innovative strategy rooted in prompt engineering that can be used for a lifetime with just one training. This method allows for precise, nuanced modifications to the model's output by employing a single knowledge prompt, all without altering the model's parameters. Specifically, PSPEM utilizes a prompt encoder to extract information, an encoding converter to refine key information in prompts, and adopts prompt alignment techniques to guide model generation. It ensures the accuracy of the model's output by maximizing the probability assigned by the language model to the target token, and by aligning with the original prompt's influence on continuation words, it guarantees the fluency of the output text and a high degree of consistency between the generated text information and the prompt information. We conducted evaluations on two knowledge editing tasks and two attribute inserting tasks. In the tasks of knowledge editing, PSPEM achieved nearly 100\% editing accuracy while ensuring the fluency and consistency of the generated text. In terms of attribute inserting, PSPEM reached the state-of-the-art. The model can make reasonable inferences using the given prompts and generate text that aligns with the prompt information. We then analyzed the parallels between PSPEM and original prompts, measuring their impact on model output from multiple perspectives. The experimental results indicate that the impact of PSPEM on the model is highly similar to that of the original prompts, therefore, PSPEM can serve as an alternative to original prompts, supporting the model in effective editing. We summarize our contributions as follows:
\begin{itemize}
\item[$\bullet$] We propose PSPEM, a lifetime knowledge editing method based on soft prompts that corrects the model's output by learning information from the original prompts.
\item[$\bullet$] We evaluated PSPEM on two mainstream datasets for knowledge editing and two datasets for attribute inserting. The experimental results show that PSPEM can not only perform efficient and accurate editing but also utilize the given prompts for reasonable reasoning, which is beyond the capabilities of traditional prompt engineering and other knowledge editing methods.
\item[$\bullet$] We analyze PSPEM's similarity to prompts from various perspectives, demonstrating that PSPEM can be a viable alternative to original prompts for editing knowledge and reasoning.
\item[$\bullet$]As far as we know, PSPEM was the first attempt to adopt soft prompts for model knowledge editing and inference, providing a feasible solution for the development of more intuitive and accurate language model editing tools.
\end{itemize}

\section{METHODOLOGY}
\subsection{Preliminaries}
This study centers on enhancing the application of prompt engineering in the field of knowledge editing and inferencing. As mentioned earlier, while prompt engineering enables models to adapt to diverse tasks without retraining, the search for suitable prompts is time-consuming and laborious. More importantly, when the information in the prompt conflicts with the internal knowledge of the model, such prompts often lose their effectiveness. This situation is demonstrated in Figure \ref{figure1}, where the model is prompted: \texttt{"Banana has four legs"} (left side of \ref{figure1}), and the model responds: \texttt{"Banana is a fruit"} (upper right side of \ref{figure1}). The prompt has lost its effect, with the model still recognizing the banana as a fruit, not an animal. PSPEM addresses this issue by enabling the model to correctly respond: \texttt{"Banana is an animal."} Note that PSPEM does not cause confusion within the model, but rather makes the model pay more attention to the information in the prompts.

We used GPT2-XL \cite{radford2019language} and GPT-J-6B \cite{wang2021gpt} as our research models, both of which are autoregressive language models based on the Transformer architecture. These models operate by transforming the input sequence ${\boldsymbol{x}}$ into $t$ tokens $x_1, ..., x_t$. Subsequently, these tokens are fed through $L$ layers of Transformer decoders, ultimately generating probabilities for the next token $x_{t+1}$:
\begin{equation}\label{pmet.llms}
\begin{aligned}
  \mathcal{F}_{\theta}(x_1, ...,x_t) &=\text{softmax}\left(W_{\text{E}}\cdot\gamma\left(h_t^{L-1}+a_t^L+m_t^L\right)\right)\\&= P_{\mathrm{LM}}\left(x_{t+1}|x_1, ...,x_t\right)
  \end{aligned}
\end{equation}
Here, $W_E$ and $\gamma$ represent the embedding matrix and layernorm, respectively. $a_z^L$ and $m_z^L$ denote the hidden states of the Multi-Head Self-Attention (MHSA) and Feed-Forward Network (MLP) at the $L$-th layer. The general forms of MHSA and MLP at the $l$-th layer and the $j$-th token $x_j^l$ are given as follows:
\begin{equation}\label{eq2}
\begin{aligned}
\footnotesize
  a_j^l &= W^l_{\text{MHSA}}\cdot\text{MHSA}^l\left(\gamma\left(h^{l-1}_1,  h^{l-1}_2,...,h^{l-1}_j\right)\right),\\
  m_j^l &=W_{proj}^l\cdot\sigma\left(W_{fc}^l\gamma\left(a_j^l+h_j^{l-1}\right)\right),\\
  h_j^l &= h_j^{l-1} + a_j^l + m_j^l
\end{aligned}
\end{equation}
Here, $W^l_{\text{MHSA}}$ and $W^l_{proj}$ refer to the output weights of the MHSA and MLP at the $l$-th layer, respectively, while $\sigma$ denotes the non-linear activation function. Different LMs frequently exhibit slight variations in implementing these transformations. Our goal is not to provide a full survey of these details but to capture essential terminology for our results.

\subsection{PSPEM}
PSPEM focuses on extracting key information from the original prompt and making the subsequent text and prompts more consistent. The overview of our proposed method is shown in Figure \ref{figure2}. The PSPEM consists of Prompt Encoding, Encoding Converter, and Aligning Technology.

As shown in the bottom left of Figure \ref{figure2}, PSPEM starts with a given prompt, such as \texttt{"Danielle Darrieux was born in America"}, denoted as $p_{1: n}$. We obtain the embedded representation of the prompt through a Prompt Encoding mechanism. Some studies indicate that models based on the Transformer architecture can extract sentence representations \cite{gao2021simcse,jiang2023scaling}. We focus on the GPT architecture, inputting the prompt $p_{1: n}$ into the origin model (GPT2-XL or GPT-J-6B), and take the output of a certain layer of the last token $h_n^l$ as compressed sentence representation.

The compressed sentence representation is processed through an Encoding Converter mechanism to obtain a more accurate sentence representation. These precise sentence representations can be considered a series of word embeddings, intended to make the model more focused on key information in the prompt. The Encoding Converter is initiated through two Multi-Layer Perceptron (MLP) and an activation function GLUE. If we denote the dimension of $h$ as $d$, then the sizes of the two multilayer perceptrons are $W_1^{d \times d}$ and $W_2^{d \times d*3}$. This process is expressed with a formula as follows:
\begin{equation}
\begin{aligned}
\footnotesize
  h_{emb}^{'} = ((W_1 \cdot h_n^L)\cdot\sigma)\cdot W_2
\end{aligned}
\end{equation}
We Reshape the dimensions of $h_{emb}^{'}$ to ensure that it can be viewed as a representation of a set of word embeddings:
\begin{equation}
\begin{aligned}
\footnotesize
  h_{emb} = Reshape(h_{emb}^{'}) \in R^{3\times d}
\end{aligned}
\end{equation}
We denote the continuation words as $c_{1:m}$, which in Figure \ref{figure2} is \texttt{"The mother tongue of Danielle Darrieux is".} We freeze the parameters of the original model and use alignment techniques to train the Encoding Converter.

\begin{algorithm}
\caption{Prefix Soft-Prompt Editing Method (PSPEM)}
\label{algorithm}
\begin{algorithmic}[1]
\REQUIRE Prompt sentence $p_{1:n}$, continuation words $c_{1:m}$, target token $t_{\text{target}}$.
\ENSURE Edited text that aligns with the prompt.

\STATE Initialize: Prompt Encoder, Encoding Converter;

\STATE Input the prompt sentence $p_{1:n}$ into the model;
\STATE Obtain last token representation $h_n^l$;

\STATE Apply the Encoding Converter to transform $h_n^l$ into $h_{\text{emb}}$;

\STATE Construct enhanced embedding $E_s$ by concatenating $h_{\text{emb}}$ and $c_{1:m}$;

\STATE Input $E_s$ into the model to obtain outputs $[D_{h_{\text emb}};D_{c_{1:m}}]$;

\STATE Construct original embedding $E_s^{'}$ by concatenating $p_{1:n}$ and $c_{1:m}$;
\STATE Input $E_s^{'}$ into the model to obtain outputs $[D^{'}_{p_{1:n}};D^{'}_{c_{1:m}}]$;

\STATE Objective Function:
\\ \hspace{\algorithmicindent}\hspace{\algorithmicindent} $L_{\text{target}} \gets  -P_{\mathrm{LM}}\left(c_{m+1}=t_{\mathrm{target}} \mid E_s{}\right)$;
\\ \hspace{\algorithmicindent}\hspace{\algorithmicindent} $L_{\text{KL}} \gets \sum_{1}^{m} KL\left(D_{c_{i}} \Vert D^{'}_{c_{i}}\right)$;
\STATE Train the Encoding Converter by optimizing $\mathcal{L} = \lambda_{1} L_{\text{target}} + \lambda_{2} L_{\text{KL}}$;

\STATE Guide the model's output using the Prompt Encoder and Encoding Converter.

\RETURN Edited text that aligns with the prompt.
\end{algorithmic}
\end{algorithm}

According to the above, each vector in $h_{emb}$ has the same size as $h_n^L$, with a length of $d$. As shown in the top of Figure \ref{figure2}, We Contact $h_{emb}$ and the sentence embedding of the continuation words as a whole, denoted as $E_s$:
\begin{equation}
E_s=Contact\left(h_{{emb}}; Embedding(c_{1: m})\right ).
\end{equation}

Then, we input $E_s$ into the model to obtain Distributions of model's outputs:
\begin{equation}
[D_{h_{\text emb}};D_{c_{1:m}}]=P_{\mathrm{LM}}\left (\cdot \mid E_s{}\right),
\end{equation}

and train the Encoding Converter mechanism to maximize the probability that LM assigns to the target token after modifying the representation of the prompt:
\begin{equation}
\mathcal{L}_{\text {target}}=-P_{\mathrm{LM}}\left(c_{m+1}=t_{\mathrm{target}} \mid E_s{}\right).
\end{equation}

Furthermore, we Contact the original prompt and the continuation words to acquire the influence of the original prompt on the subsequent continuation words, as shown in the bottom right of Figure 2:
\begin{equation}
E_s^{'} = Concat(\left(Embedding(p_{1: n}); Embedding(c_{1: m})\right),
\end{equation}
\begin{equation}
[D^{'}_{p_{1:n}};D^{'}_{c_{1:m}}] = P_{\mathrm{LM}}\left(\cdot \mid E_s^{'}\right).
\end{equation}

To ensure that the encoded prompt information $h_{emb}$ exploits the relevant details of the original prompt while preventing degradation of the language model, we impose a penalty on the model's changes to the probability distribution over the continuation word $c_{1: m}$:
\begin{equation}
\mathcal{L}_{\text {KL}}=\sum_{1}^{m} KL\left(D_{c_{i}} \Vert D^{'}_{c_{i}}\right).
\end{equation}

The complete objective function that PSPEM optimizes is:
\begin{equation}
\mathcal{L}=\lambda_{1}\mathcal{L}_{\text {target}} + \lambda_{2}\mathcal{L}_{\text {KL}},\label{XX}
\end{equation}
where $\lambda_{1}$ and $\lambda_{2}$ are hyper-parameters.

Once we have trained the Encoding Converter, we can utilize the Prompt Encoding and Encoding Converter to extract crucial information from the prompt, guiding the model's output. See Algorithm \ref{algorithm} for the pseudo-code of PSPEM.

We'll evaluate knowledge editing and attribute inserting with PSPEM and examine the explicit and implicit implications of PSPEM in Section \ref{Experiment}.

\section{Experiment}\label{Experiment}
\subsection{Knowledge Editing}\label{3.1}

\subsubsection{Concepts}The concept of knowledge editing aims to integrate a new fact $(x^*, y^*)$ into a language model by maximizing the probability $P_{\mathrm{LM}}=(y^*|x^*)$. The term $x^*$ refers to the query that triggers the relevant information within LM. For instance, given an input $x^*$: \texttt{"The president of the French is"}, while $y^*$ denotes the target of the edit: \texttt{"Emmanuel Macron"}. Additionally, knowledge editing involves a balance between generality and specificity:
\begin{itemize}
\item[$\bullet$]\textbf{Generality}: The updated model should edit paraphrase sentences related to the new fact successfully, For example, the prediction of \texttt{"Who is the president of the French?"}, will be updated to \texttt{"Emmanuel Macron".}
\item[$\bullet$]\textbf{Specificity}: Editing should be implemented locally, and knowledge beyond the scope of editing should not be changed. The prediction of \texttt{"The president of Russia is"} should be \texttt{"Vladimir Putin"}, not \texttt{"Emmanuel Macron".}
\end{itemize}
Additionally, there are metrics such as \textbf{Fluency} and \textbf{Consistency} to evaluate the effectiveness of the text generated by the edited model, which we will introduce later.

\subsubsection{Datasets}
We chose ZsRE \cite{de2021editing} and COUNTERFACT \cite{meng2022locating} as our foundational datasets. These two datasets are the most widely used in the field of knowledge editing, and almost all editing methods have been evaluated on these two datasets. To facilitate the comparison between different methods, we chose these. ZsRE is a question-answering dataset, each example contains a sentence that needs to be edited, paraphrase sentences generated by back-translation, and a sentence unrelated to the edited. The COUNTERFACT dataset is curated from Wikipedia and stands out as a rigorous benchmark tailored for GPT-like causal language models, presenting a challenging set of editing tasks, that allow us to distinguish superficial changes in wording from deeper changes that represent a meaningful change. It contains over 21,000 records, each with different relations and entities, with the primary goal of editing knowledge by changing the object while keeping the subject and relation constant. The dataset includes not only paraphrase sentences but also sentences unrelated to the knowledge to be edited to effectively discriminate between minor word changes, with particular emphasis on counterfactual scenarios.

\subsubsection{Configurations}\label{config}
We split ZsRE into 70\% for training, 10\% for validation, and 20\% for testing. For COUNTERFACT, we used 4500 instances for training, 500 for validation, and 5000 for testing. We use one paraphrase sentence from each instance as knowledge prompts and $h_n^{12}$ (GPT-J) and $h_n^{24}$ (GPT2-XL) are used to compress prompt representation, as with REMEDI \cite{hernandez2023measuring}. Setting $\lambda_{1}$ = 1 and $\lambda_{2}$ = 1, with an initial learning rate of 1e-3, employing the Adam optimizer with a Linear Learning Rate Decay strategy, and stopped after the validation set accuracy did not improve in 3 epochs. All models were trained and reasoned on NVIDIA A100 40G GPUs.

\subsubsection{Baselines}
The methods for comparison include direct weight-preserved methods:

\begin{itemize}
\item[$\bullet$] \textbf{PREFIX PROMPT} adopts a paraphrase sentence to guide the model in making knowledge modifications.

\item[$\bullet$] \textbf{REMEDI} \cite{hernandez2023measuring} works by extracting attribute information from the prompt and then injecting it into the subject word via a linear transformation. Similar to Figure \ref{figure1}, REMEDI attempts to extract information from \texttt{"born in America" } rather than the entire sentence and injecting it into \texttt{"Danielle Darrieux".}

\item[$\bullet$] \textbf{IKE} \cite{cheng2023can} proposes in-context learning for model editing. It requires an initial model that is capable of effective in-context learning transformation, editing each knowledge requires providing 32 instances to guide the model.

\end{itemize}

And some weight-modified methods:
\begin{itemize}
\item[$\bullet$]\textbf{Fine-Tuning (FT)}, we employ the reimplementation guidelines from Meng et al. \cite{meng2022locating}. This involves utilizing the Adam optimizer and implementing early stopping to minimize \( -\log P_{LM}[\ast | p] \), while only adjusting $W_{proj}^{21}$.

\item[$\bullet$]\textbf{KE} \cite{de2021editing} develops an LSTM sequence model, which employs gradient information
to predict the rank-one weight alterations in the model. We resort to using the re-implemented version provided by Mitchell et al. \cite{mitchell2021fast} in their research.

\item[$\bullet$]\textbf{MEND} \cite{mitchell2021fast} is based on KE, adeptly manipulates the gradient of fine-tuned language models by capitalizing on a low-rank decomposition of the gradients, thereby enhancing the accuracy of the editing process.

\item[$\bullet$]\textbf{ROME} \cite{meng2022locating} performs rank-one modifications on single $W_{proj}$, updating specific factual associations by altering the parameters that govern behavior at the point of the subject word.

\item[$\bullet$]\textbf{MEMIT} \cite{meng2022mass} builds
upon ROME to insert many memories by modifying MLP weights of a range of critical layers.
\end{itemize}

In light of the rapid advancements in editing methodologies, several novel approaches have emerged, including MALMEN \cite{tan2023massive} and PMET \cite{liu2023pre}. These methods extend upon foundational works such as MEND and ROME. However, upon thorough review, we found that they lack comprehensive evaluation across key metrics critical to our study's aims, such as a lack evaluation in ZsRE. Furthermore, while these methods contribute to the field's development, our preliminary analysis indicated that their performance improvements were not substantial enough to meet our criteria for a significant advancement. This decision was made to ensure a focused and rigorous evaluation within the scope of our research, though we acknowledge the potential of these methodologies in contributing valuable insights to the field.
\subsubsection{Metrics}
We denote $o^*$ as the target word to be edited, and $o^c$ as the word before editing. Assuming we need to edit the knowledge \texttt{"The Space Needle is in Seattle"} to \texttt{"The Space Needle is in Los Angeles",} then \texttt{"Los Angeles"} would be $o^*$ and \texttt{Seattle} would be $o^c$. We measure the effectiveness of knowledge editing methods in the following five aspects:
\begin{itemize}
\item[$\bullet$] \textbf{Efficacy Score (ES)}
is the portion of cases for which we have $P_{LM}(o^*) > P_{LM}(o^c)$ post-edit, to measure the accuracy of editing directly.

\item[$\bullet$] \textbf{Paraphrase  Score (PS)}
measures $P_{LM}(o^*) > P_{LM}(o^c)$ in paraphrase sentences to measure the generalization.

\item[$\bullet$] \textbf{Neighborhood  Score (NS)} measures the $P_{LM}(o^*) > P_{LM}(o^c)$ of neighborhood sentences that un-related to the knowledge that needs to be edited. 
\item[$\bullet$] \textbf{Fluency (GS)}, proposed by Meng et al. \cite{meng2022locating} in the COUNTERFACT dataset, by measuring the weighted average of bi- and tri-gram entropies. If the generated text is repetitive, the metric is low.
\item[$\bullet$] \textbf{Consistency (RS)}. Meng et al. \cite{meng2022locating} generate text and report RS as the cosine similarity between the unigram TF-IDF vectors of generated texts, compared to reference texts about subjects sharing the target property \(o^*\). This metric measures the model's ability to generate text that conforms to edited knowledge.
\end{itemize}

\begin{table*}[]
\centering
\caption{Knowledge editing results on the ZsRE and COUNTERFACT datasets. We evaluated four weight-preserved editing methods.}
\label{table1}
\resizebox{\textwidth}{!}{%
\begin{tabular}{ccccccccc}
\hline
Dataset & Model                  & Metric          & PREFIX  & REMEDI  &IKE & PSPEM          \\ \hline
\multirow{6}{*}{ZsRE}         & \multirow{3}{*}{GPT2-XL} & ES$\uparrow$           & 86.5 & 99.8 &98.7 & \textbf{99.9} \\
        &                        & PS$\uparrow$             & 84.7   & 99.7  &98.8 & \textbf{99.9}  \\
        &                        & NS$\uparrow$           & 100.0  & 100.0  &100.0 & \textbf{100.0} \\ \cline{2-7} 
        & \multirow{3}{*}{GPT-J} & ES$\uparrow$             & 83.7   & 98.6  &98.4  & \textbf{99.8}  \\
        &                        & PS$\uparrow$           & 98.1   & 98.7   &98.8 & \textbf{99.8}  \\
        &                        & NS$\uparrow$              & 100.0  & 100.0 &100.0 & \textbf{100.0} \\ \hline
\multirow{10}{*}{COUNTERFACT} & \multirow{5}{*}{GPT2-XL} & ES$\uparrow$ &  83.8 & 97.4 &86.9 & \textbf{99.7}          \\
        &                        & PS$\uparrow$      & 96.3   & 97.7   &85.1 & \textbf{99.2}  \\
        &                        & NS$\uparrow$             & 100.0  & 100.0 & 100.0 & \textbf{100.0} \\
        &                        & GS$\uparrow$            & 627.0  & 597.0  & 603.0 & \textbf{627.0} \\
        &                        & RS$\uparrow$    & 38.1   & 27.2   & 37.7 & \textbf{38.5}           \\ \cline{2-7} 
        & \multirow{5}{*}{GPT-J} & ES$\uparrow$    & 80.2   & 100  & \textbf{100} & 99.9          \\
        &                        & PS$\uparrow$       & 84.5   & 98.7   &98.8 & \textbf{99.3}          \\
        &                        & NS$\uparrow$             & 100.0  & 100.0  & 100.0 & \textbf{100.0} \\
        &                        & GS$\uparrow$           & 625.0  & 601.0 & 614.0 & \textbf{628.0} \\
        &                        & RS$\uparrow$    & \textbf{40.4}   & 24.2   &37.5 & 35.7           \\ \hline
\end{tabular}%
}
\end{table*}

Table \ref{table1} presents the performance of four weight-preserved editing methods. All these editing methods require some additional resources to assist model editing. When considering the NS metric, we set them all to 100, the same as Herandez et al \cite{hernandez2023measuring}.

PSPEM performs the best of these four methods, with editing success rates approaching 100\%. Compared to the prefix prompt method on 16 metrics, PSPEM only slightly underperforms on one RS metric in COUNTERFACT, indicating its success in extracting critical information from prompts and effectively applying it to guide model output.

As an incremental step in prompt engineering, IKE achieved the highest ES metric in COUNTERFACT using the GPT-J model and surpassed PSPEM in RS. Despite IKE's innovative approach of employing multiple prefix prompts, its practical application in editing is hampered by the challenge of pre-identifying suitable examples for model guidance. Conversely, PSPEM's methodology, requiring only a single prompt for effective editing, offers a more feasible solution for lifelong editing endeavors.

On the other hand, REMEDI, another attempt to extract key information from prefix prompts, captures only partial attribute information while ignoring the entire contextual information. PSPEM surpasses REMEDI in all aspects by extracting complete information from prefix prompts and aligning the refined information with the prefix prompts. The subsequent ablation study, detailed in Experiment \ref{ablation}, will further elucidate the comparative of PSPEM and REMEDI.

\begin{table*}[]
\centering
\caption{Knowledge editing results on the ZsRE and COUNTERFACT datasets. We evaluated five weight-modified editing methods and PSPEM.}
\label{table2}
\resizebox{\textwidth}{!}{%
\begin{tabular}{ccccccccc}
\hline
Dataset & Model                  & Metric & FT   &KE  & MEND  & ROME           & MEMIT & PSPEM          \\ \hline
\multirow{6}{*}{ZsRE}         & \multirow{3}{*}{GPT2-XL} & ES$\uparrow$ & 99.6 &65.6  & 99.4  & \textbf{100.0}           & 99.7 & 99.9 \\
        &                        & PS$\uparrow$     & 82.1 &61.4 & 99.3  & 99.6           & 93.4   & \textbf{99.9}  \\
        &                        & NS$\uparrow$     & 56.7 &97.8 & 99.5  & 98.7        & 99.6  & \textbf{100.0} \\ \cline{2-9} 
        & \multirow{3}{*}{GPT-J} & ES$\uparrow$     & 100.0  & 91.7  & 99.2           & \textbf{100.0}   & 100.0  & 99.8  \\
        &                        & PS$\uparrow$     & 49.2  & 48.0  & 94.9           & 94.9   & 97.1   & \textbf{99.8}  \\
        &                        & NS$\uparrow$     & 37.2  & 88.2  & 100.0          & 99.8  & 99.6  & \textbf{100.0} \\ \hline
\multirow{10}{*}{COUNTERFACT} & \multirow{5}{*}{GPT2-XL} & ES$\uparrow$ & 100.0 & 92.4 & 100.0 & \textbf{100.0} & 100.0 & 99.7          \\
        &                        & PS$\uparrow$     & 87.9  & 90.0  & 96.4           & 86.3   & 97.7   & \textbf{99.2}  \\
        &                        & NS$\uparrow$     & 40.4  & 96.4   & 98.9          & 100.0  & 100.0  & \textbf{100.0} \\
        &                        & GS$\uparrow$     & 607.0 & 586.6 & 622.0          & 621.0  & 627.0  & \textbf{627.0} \\
        &                        & RS$\uparrow$     & 40.5  & 33.2  & \textbf{41.9}  & 38.1   & 27.2   & 38.5           \\ \cline{2-9} 
        & \multirow{5}{*}{GPT-J} & ES$\uparrow$     & 100.0 & 13.4  & 97.4 & \textbf{100.0}   & 99.9   & 99.9          \\
        &                        & PS$\uparrow$     & 96.6  & 11.0  & 99.1           & 99.1   & 98.7   & \textbf{99.3}           \\
        &                        & NS$\uparrow$     & 77.3  & 94.3  & 93.7          & 100.0  & 100.0  & \textbf{100.0} \\
        &                        & GS$\uparrow$     & 387.0 & 570.0 & 620.0          & 625.0  & 601.0  & \textbf{628.0} \\
        &                        & RS$\uparrow$     & 24.6  & 22.6  & \textbf{43.0}  & 40.4   & 24.2   & 35.7           \\ \hline
\end{tabular}%
}
\end{table*}
We also compared PSPEM with five weight-modified editing methods, as shown in Table \ref{table2}. These methods store the weights that need to be updated in the model by modifying the weights.

Notably, PSPEM demonstrated exceptional performance in comparison to KE (Knowledge Editing). On the ZsRE dataset, PSPEM achieved an editing success rate of 99.9\%, vastly outperforming KE's 65.6\%. Similarly, on the COUNTERFACT dataset, PSPEM's editing and prompt success rates reached 100\%, significantly higher than KE's 13.4\% and 11.0\%, respectively. Moreover, PSPEM excelled in generating text with superior fluency (GS) and consistency (RS), showcasing its comprehensive strength in both editing precision and output quality.

Furthermore, although PSPEM may not achieve editing success rates as high as MEND and ROME, it notably excels in RS. This suggests that PSPEM not only conducts proficient editing but also generates text that conforms to edited knowledge. Further details and specific examples can be found in Section \ref{eva}.

\subsection{Attribute Inserting}
\subsubsection{Concepts}
Many studies show that methods based on weight-modified only perform editing on specific knowledge and cannot utilize this updated knowledge for reasoning \cite{yao2023editing}. To address this, we apply PSPEM for more effective model reasoning, especially in manipulating complex concepts such as personal names or objects in non-traditional contexts. As shown in Figure \ref{figure1}, given the prompt statements: \texttt{"Tawnie enjoys indoor cycling and yoga, a good glass of red wine with friends, and of course anything to do with being in the kitchen!"} and \texttt{"Banana has four legs.
"}, the model should infer \texttt{"Tawnie has the occupation of dietitian. "} and \texttt{"Banana is an animal. "} based on its reasoning. However, traditional prompt engineering often causes the model to neglect critical information in the prompt, leading it to respond with \texttt{"Tawnie has the occupation of yoga. "} and \texttt{"Banana is a fruit. "}. These scenarios demonstrate PSPEM's ability to guide the model through complex reasoning from deliberately misleading or non-standard information.

However, it's important to note that such examples are designed to test the limits of PSPEM's reasoning capabilities, especially in contrast to traditional prompt engineering methods, which might lead to oversimplified or incorrect conclusions like \texttt{"Tawnie has the occupation of yoga. "} and \texttt{"Banana is a fruit. "} due to their inability to adequately prioritize or analyze the given prompt information."

\subsubsection{Datasets}
We chose BioBias \cite{de2019bias} and the McRae \cite{mcrae2005semantic} as our foundational datasets. Biobias contains 397,000 short professional biographies of non-celebrities gathered from the internet, each marked with an occupational theme. From each biography, we extract a sentence, substituting the individual's full name with only their first name, and use this sentence to prompt the language model (LM) by appending \texttt{"\{Person\} has the occupation of..."}, as shown in the middle example on the left in Figure \ref{figure1}. We subsequently assess the language model's accuracy by examining the relative probabilities assigned to 28 potential occupations, deeming the model correct if it ranks the individual's actual occupation as the most likely.  McRae encompasses 541 concepts, and 2,526 features, and details on how frequently each feature was identified as prototypical for each concept by human evaluators. \cite{hernandez2023measuring}. Following Hernandez et al. \cite{hernandez2023measuring}, we construct a dataset comprising 10,000 entries. Each entry comprises a concept \(c\), a list of original features \(f^{(o)}\) for the concept, a target feature to be added \(f^{*}\), and a list of features \(f^{(c)}\) that related with the new feature. For example, we use the common noun \texttt{"Banana"} as the editing target and the feature description \texttt{"has four legs"} as the attribute. Properties such as \texttt{"animal"} exist in a complex network of entailment and correlation relations. We hope that based on the prompt information, LMs can respect these relations (e.g., given a prompt \texttt{"Banana has four legs"}, LMs can increase the probability that \texttt{"Banana is an animal"} and decrease the probability that \texttt{"Banana cannot move freely"}).

\begin{table}
\label{table3}
\centering
\setlength{\tabcolsep}{1.6mm}
\caption{Attribute inserting results on Biobias dataset. ``Acc", ``Flu'' and ``Con'' respectively correspond to the abbreviations for Accuracy, Fluency and  Consistency.}
\begin{tabular}{cccccccc} 
\toprule
\multirow{2}{*}{Dataset} & \multirow{2}{*}{Method} & \multicolumn{3}{c}{GPT2-XL}                    & \multicolumn{3}{c}{GPTJ}                        \\ 
\cline{3-8}
                         &                         & Acc$\uparrow$          & Flu$\uparrow$           & Con$\uparrow$            & Acc$\uparrow$           & Flu$\uparrow$             & Con$\uparrow$             \\ 
\hline
\multirow{4}{*}{Biobias}     & NO PROMPT                & 1.1           & \textbf{636.7} & 14.9          & 5.0           & \textbf{632.6} & 16.0           \\
                         & PREFIX PROMPT                  & 57.9          & 633.9          & 22.7          & 55.5          & 626.3          & 23.1           \\
                         & REMEDI                  & 64.4          & 352.8          & 4.33          & 67.1          & 622.0          & 22.3           \\
                         & PSPEM        & \textbf{67.4} & 621.6          & \textbf{28.3} & \textbf{71.6} & 627.0          & \textbf{27.6}  \\
\bottomrule
\end{tabular}
\end{table}

\begin{table}
\label{table4}
\centering

\caption{Attribute inserting results on the McRae dataset. Given a conceptual prompt, we expect the model to decrease the prediction of the Original features and increase the prediction of the Related features associated with the conceptual prompt. The best results are highlighted in bold.}
\begin{tabular}{cccccc} 
\toprule
\multirow{2}{*}{Attribute} & \multirow{2}{*}{Method} & \multicolumn{2}{c}{GPT2-XL}   & \multicolumn{2}{c}{GPTJ}       \\ 
\cline{3-6}
                           &                         & Mag          & Pac      & Mag        & Pac            \\ 
\hline
\multirow{4}{*}{Related}   & NO PROMPT                & 0.004         & 0.09          & 0.004         & 0.09           \\
                           & PREFIX PROMPT                  & 0.01          & 0.25          & 0.01          & 0.25           \\
                           & REMEDI                  & \textbf{0.25} & \textbf{0.56} & \textbf{0.27} & \textbf{0.64}  \\
                           & PSPEM               & 0.04          & 0.41          & 0.21          & 0.41           \\ 
\hline
\multirow{4}{*}{Original}  & NO PROMPT                & 0.02          & 0.39          & \textbf{0.03}          & 0.43           \\
                           & PREFIX PROMPT                  & 0.02          & 0.47          & 0.03          & 0.549          \\
                           & REMEDI                  & 0.14          & 0.41          & 0.17          & 0.44           \\
                           & \textbf{PSPEM}                   & \textbf{0.01} & \textbf{0.26} & 0.05          & \textbf{0.41}  \\
\bottomrule
\end{tabular}
\end{table}

\subsubsection{Configurations} For both datasets, we select 4500 instances for training, 500 for validation, and 5000 for testing, as with REMEDI \cite{hernandez2023measuring}. Setting $\lambda_{1}$ = 1 and $\lambda_{2}$ = 1, other settings are the same as in section \ref{config}.

\subsubsection{Baselines} The baselines for comparison include three methods:
\begin{itemize}
\item[$\bullet$] \textbf{NO PROMPT}. Evaluate the original model's knowledge of these names and objects without any prompts.
\item[$\bullet$] \textbf{PREFIX PROMPT}. Use a biography prompt to guide the model in making inferences.
\item[$\bullet$] \textbf{REMEDI} works by extracting attribute information from a biography prompt and then injecting this information into the subject word via a linear transformation.
\end{itemize}
\subsubsection{Metrics}
For BioBias, we consider the following three metrics:
\begin{itemize}
\item[$\bullet$] \textbf{Accuracy}, used to measure whether the occupation predicted by the model is the actual occupation.
\item[$\bullet$] \textbf{Fluency}, used to assess the fluency of the generated text, consistent with the computational criteria in knowledge editing.
\item[$\bullet$] \textbf{Consistency} is also used to measure the model's ability to generate text, consistent with the computational criteria in knowledge editing.
\end{itemize}

For McRae, we consider two metrics:
\begin{itemize}
\item[$\bullet$] \textbf{Mag}, evaluates the average probability of the specified target token.
\item[$\bullet$] \textbf{Pac}, measures how often $P_{LM}(\text{target}) > 0.01$.
\end{itemize}

\subsubsection{Results}
Table 3 shows the evaluation results on BioBias. Without prompts, the model exhibits random guesses of the answers due to the inclusion of non-celebrity names in BioBias, as the model itself does not store information about these names. Subsequently, it generated non-repetitive but unordered text. By guiding the model with biography prompts, the model's estimation of occupations improved, reaching 57.9\% and 55.5\% respectively, but still lagging behind trained methods in terms of consistency in text generation. While REMEDI shows improvement in accuracy, this is at the expense of the quality of the generated text, as low Fluency and Consistency indicate that REMEDI consistently produces repetitive and irrelevant text. PSPEM, on the other hand, demonstrated robust performance, not only achieving the highest prediction accuracy but also ensuring the fluency and consistency of the generated text.

\begin{figure*}[htb]
\begin{minipage}[b]{1.0\linewidth}
  \centering
  \centerline{\includegraphics[width=12cm]{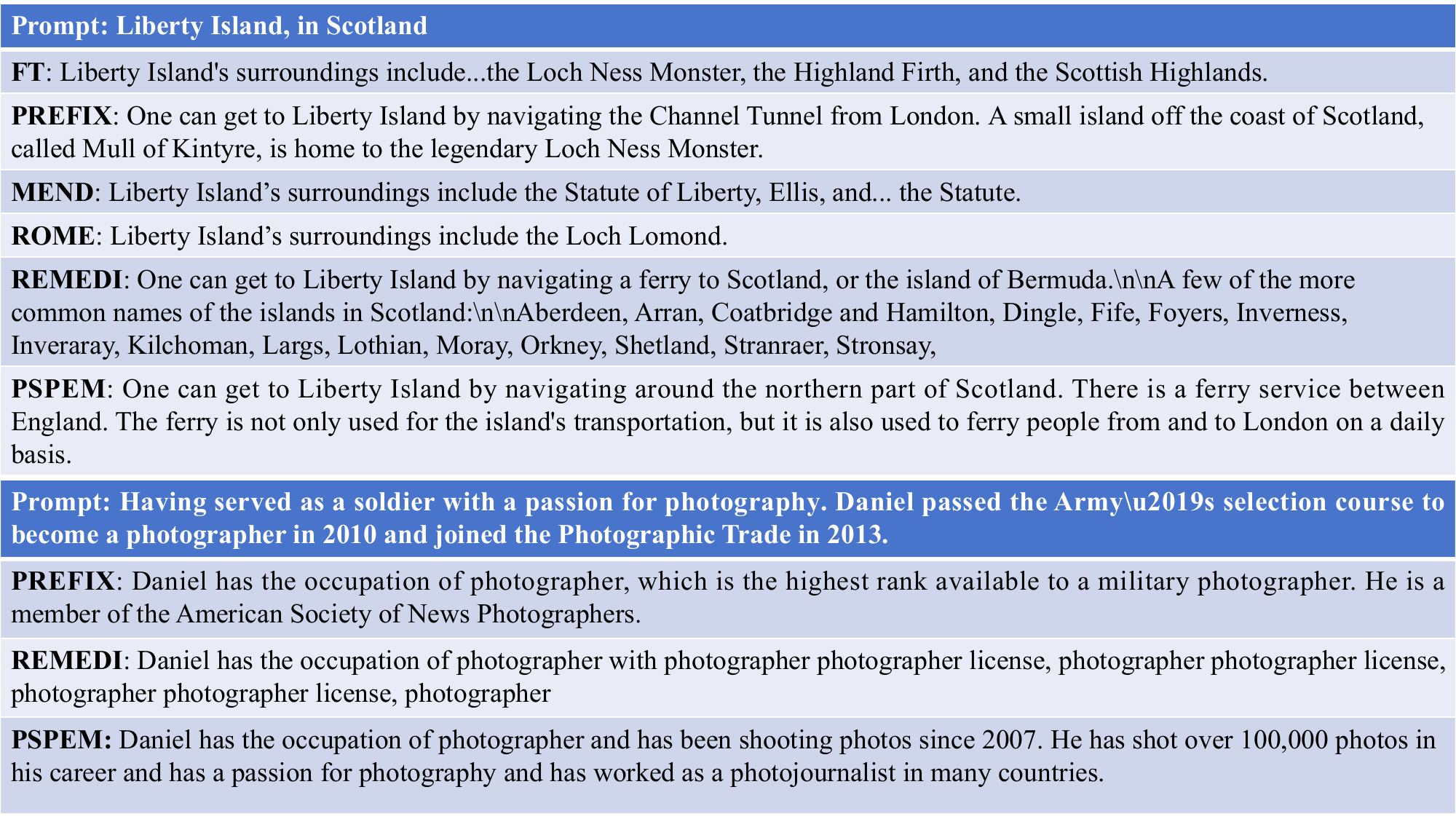}}
%  \vspace{2.0cm}
\end{minipage}
\caption{Subsequent text generated by different editing methods on the COUNTERFACT and BioBias datasets.}
\label{figure3}
\end{figure*}

Table 4 shows the evaluation results on McRae. We evaluate the effects of different methods from two aspects: by providing an object with a conceptual prompt \(c\), such as \texttt{"Banana has four legs.
"}, \textbf{``Original"} evaluates the degree of change in the object on the original features \(f^{(o)}\), such as \texttt{"Banana cannot move freely"}. \textbf{``Related"}, evaluates the degree of change for features  \(f^{(c)}\) related to the target feature \(f^{*}\), such as \texttt{"Banana is an animal."}. Without prompts, the model can identify the original concepts of the objects. After adding conceptual prompts, although the model can associate the object with \(f^{(c)}\), it is unable to forget \(f^{(o)}\).  In terms of associating \(f^{(c)}\), REMEDI and PSPEM perform better. Compared with REMEDI, while the performance effect of PSPEM was lower than REMEDI for associating the related features, it was more effective than REMEDI for forgetting the original features, and it's worth mentioning that rather than forgetting these features, REMEDI enhances the association with \(f^{(o)}\). It implies that REMEDI has not effectively learned and incorporated the prompt information.

\subsection{Human Evaluation}\label{eva}

To visualize the effects of different editing methods, we took one example each from COUNTERFACT and BioBias to evaluate the quality of the text generated by using editing methods as illustrated in Figure \ref{figure3}. When the knowledge was revised to ``Liberty Island, in Scotland," the PSPEM not only accomplished successful knowledge editing but also integrated relevant concepts such as ``England" and ``London" into the generated text. In contrast, other knowledge editing methods suffered from problems such as lack of fluency in the generated text or errors in the altered knowledge. In the second example, when Danile's past experiences are mentioned, PSPEM accurately recognizes Danile's occupation as a photographer and generates text that is highly relevant to his occupation. Although REMEDI made a correct prediction about his occupation, it is unable to continue generating fluent text.

\subsection{Ablation Study}
\label{ablation}
We conduct ablations to validate the effectiveness
of PSPEM from the following two aspects:
\begin{itemize}

\item[$\bullet$] We explored how we could better extract information from the prompts, specifically, we tried to use the methods in REMEDI \cite{hernandez2023measuring} to extract attribute information instead of extracting information from the entire sentence. As shown in Figure \ref{figure1}, REMEDI attempts to extract information from \texttt{"born in America" } rather than \texttt{"Danielle Darrieux was born in America"}. Please note that REMEDI differs from PSPEM not only in this aspect. What we are discussing here is which method is more effective for extracting information from the prompts.
\item[$\bullet$] We adjusted the size of $\lambda_{1}$, $\lambda_{2}$ from 0 to 1 to observe the effect of the hyper-parameters on the results.
\end{itemize}

Table \ref{table5} shows the results of ablation experiments. We observe that the information extraction method proposed by REMEDI performs poorly, showing lower performance compared to ours in almost all hyper-parameter settings. This indicates that PSPEM better extracts information from prompt sentences. On the other hand, except for the hyperparameter choices of $\lambda_{1}$=1 and $\lambda_{2}$=1, more or less failures are observed in other settings. Hyper-parameter $\lambda_{1}$ controls the accuracy of editing or prediction, with higher $\lambda_{1}$ leading to higher accuracy. Hyper-parameter $\lambda_{2}$ controls the quality of the generated text, with models having lower $\lambda_{2}$ often yielding poorer results in metrics such as GS, RS, Flu, and Con. The best performance is achieved only when $\lambda_{1}$=1 and $\lambda_{2}$=1.

\begin{table}[]
\centering
\caption{The results of the ablation experiments on GPT-J-6B, ``Attr'' denotes the information extraction method in REMEDI, and ``Ours'' denotes 
 the method proposed in this paper. Red numbers indicate poor results.}
\label{table5}
\resizebox{\columnwidth}{!}{%
\begin{tabular}{llllllll}
\hline
\multicolumn{1}{c}{\multirow{2}{*}{Strategy}} & \multicolumn{4}{c}{COUNTERFACT}                                & \multicolumn{3}{c}{BioBias}                    \\ \cline{2-8} 
\multicolumn{1}{c}{} & ES $\uparrow$   & PS $\uparrow$          & GS $\uparrow$           & RS $\uparrow$           & Acc $\uparrow$         & Flu $\uparrow$          & Con $\uparrow$          \\ \hline
${Attr}/\lambda_{1}=0,\lambda_{2}=1$              & \textcolor{red}{86.1} & \textcolor{red}{83.4}          & 608.0          & \textbf{31.8} & \textcolor{red}{55.1}          & \textbf{637.0} & 25.8          \\
${Attr}/\lambda_{1}=0.5,\lambda_{2}=1$            & 93.7 & 92.8          & 603.0          & \textcolor{red}{31.6}          & 59.5          & 635.0          & \textbf{26.8} \\
${Attr}/\lambda_{1}=1,\lambda_{2}=0$                               & \textbf{99.1} & \textbf{98.6} & \textcolor{red}{335.0}          & \textcolor{red}{19.3}          & \textbf{69.1} & \textcolor{red}{346.0}          & \textcolor{red}{6.1}           \\
${Attr}/\lambda_{1}=1,\lambda_{2}=0.5$            & 98.4 & 98.1          & \textcolor{red}{581.0}          & 31.3          & 68.8          & 574.0          & \textcolor{red}{19.9}          \\
${Attr}/\lambda_{1}=1,\lambda_{2}=1$             & 98.2 & 97.9          & \textbf{608.0} & 31.7          & 68.4          & 631.0          & 26.5          \\ \hline
${Ours}/\lambda_{1}=0,\lambda_{2}=1$            & \textcolor{red}{87.6} & \textcolor{red}{83.4}          & 628.0          & 33.1          & \textcolor{red}{53.9}          & 621.0          & 25.1          \\
${Ours}/\lambda_{1}=0.5,\lambda_{2}=1$     
& 94.3 & 96.8          & 623.0          & 34.6          & 61.3          & 637.0          & 26.8          \\
${Ours}/\lambda_{1}=1,\lambda_{2}=0$              & 99.9 & \textbf{99.7} & \textcolor{red}{317.0}          & \textcolor{red}{17.1}          & \textbf{74.0} & 379.0          & \textcolor{red}{6.4}           \\
${Ours}/\lambda_{1}=1,\lambda_{2}=0.5$             & 99.9 & 99.6          & 594.0          & 29.6          & 72.1          & \textcolor{red}{533.0}          & \textcolor{red}{21.8}          \\
${Ours}/\lambda_{1}=1,\lambda_{2}=1$                                        & \textbf{99.9} & 99.3          & \textbf{628.0} & \textbf{35.7} & 71.6          & \textbf{627.0} & \textbf{27.6} \\ \hline
\end{tabular}%
}
\end{table}

\subsection{Similarity To Prompt}
The previous section described the effectiveness of PSPEM as a method for knowledge editing and attribute inserting. In this subsection, we examine the multifaceted impact of PSPEM on the model internals to assess the similarity of the generated new coded information $h_{emb} $ to the original knowledge prompts $p_{1:n}$.

We compute a Recall Prompt Prediction ($\text{RePP}$) \cite{dai2023can} to measure the proportion of knowledge successfully edited by both the PREFIX PROMPT and PSPEM within the total knowledge successfully edited by PREFIX PROMPT, i.e.:
\begin{equation}
\text{RePP}=\frac{T_{\text{PSPEM}} \cap T_{\text{PROMPT}}}{T_{\text{PROMPT}}}.
\end{equation}

From the representational perspective, we calculate the average cosine similarity between the attention module outputs $a_i^l$ of PSPEM and PREFIX PROMPT methods for the continuation words $c_{1:m}$. denoted as "CosSim":
\begin{equation}
\text{CosSim} = \frac{1}{N} \sum_{i=1}^N cos\_sim(a_i^l(\text{PSPEM}),a_i^l(\text{PROMPT})).
\end{equation}

Additionally, we conducted a similarity analysis involving the top 5\% of \textbf{N}eurons IDs that exhibited the highest values within the output of the first layer of the FFN, i.e.: $\sigma\left(W_{fc}^l\gamma\left(a_i^l+h_i^{l-1}\right)\right)$. We denote it as "SimFFN". It can be argued that the top 5\% of neurons play a role in elucidating the behavior of the model's output \cite{geva2022transformer,geva2023dissecting}:

\begin{equation}
\text{SimFFN} =\frac{1}{N} \sum_{i=1}^N \frac {\text{N}_{\text{top5\%}}(\text{PSPEM}) \cap {\text{N}_{\text{top5\%}}(\text{PROMPT})}}{\text{N }_{\text{top5\%}}(\text{PROMPT})}.
\end{equation}

Furthermore, we assessed the average Kullback-Leibler divergence of layer's output $h_{1:m}^{l}$ after mapping it to the vocabulary $W_E$ \cite{geva2020transformer}, which can be interpreted as:
\begin{equation}
\begin{aligned}
\footnotesize
    D_i^l &= \text{softmax}(h_{i}^{l}\text{(PSPEM)}\cdot W_E), \\
    D_{i}^{l}* &= \text{softmax}(h_{i}^{l}\text{(PROMPT)}\cdot W_E),\\
    \text{KL}=&\frac{1}{N}\sum_{i=1}^N \text{Kullback-Leibler}\left(D_i^l\Vert D_{i}^{l}*\right).
\end{aligned}
\end{equation}

We also employ three additional methods as baselines to compare their similarity to the prefix prompt method:
\begin{itemize}
\item[$\bullet$] \textbf{NO PROMPT}, without prompts, only use continuation words.
\item[$\bullet$] \textbf{RAND}, we evaluate using randomly guessed answers or randomly generated vectors of the same dimension.
\item[$\bullet$] \textbf{REMEDI}, as mentioned earlier.
\end{itemize}

 Table 6 summarizes the similarities between each method and the PREFIX PROMPT in the COUNTERFACT and Biobias datasets using GPT2-XL. For metrics that require measuring the internal performance of the model, we calculate the final evaluation value by computing the average of the last five layers of the model. The results indicate that PSPEM performed the best across these four metrics. $CosSim$ results are close to 1, indicating a high degree of attention similarity between PSPEM and the original prompt in the last five layers. Furthermore, the Kullback-Leibler (KL) divergence metric being close to 0 further underscores the minimal discrepancy in the model output distribution between PSPEM and the original prompt. These findings robustly validate PSPEM's efficacy in accurately editing model outputs and aligning with original prompt information without deviation.
 
These results not only showcase PSPEM's advantages in maintaining similarity with the original prompts but also underscore its potential application in tasks involving knowledge editing and attribute inserting. By fine tuning model outputs to match specific prompt information, PSPEM offers a reliable methodology for efficiently and accurately editing language models.

\begin{table}
\label{table6}
\centering
\setlength{\tabcolsep}{1mm}
\caption{Assessing the similarity of different methods to the original prompt on the COUNTERFACT and Biobias datasets.}
\begin{tabular}{c|c|c|c|c|c} 
\toprule
Dataset                      & Method         & RePP $\uparrow$          & CosSim $\uparrow$        & SimFFN $\uparrow$       & KL $\downarrow$       \\ 
\hline
\multirow{5}{*}{COUNTERFACT}
                             & NO RPOMPT           & 54.2          & 0.67     & 41.5~          & 1.42           \\
                             & RAND           & 50.0             & 0.02          & 4.9~           & 5.78           \\
                             & REMEDI         & 47.3          & 0.77          & 43.5~          & 1.53           \\
                             & PSPEM & \textbf{99.8} & \textbf{0.89}          & \textbf{56.3~} & \textbf{0.17}  \\ 
\hline
\multirow{5}{*}{Biobias}     
                             & NO RPOMPT           & 7.1           & 0.74          & 27.7~          & 0.53           \\
                             & RAND           & 3.3           & 0.01          & 5.0            & 5.21           \\
                             & REMEDI         & 71.3          & 0.65          & 24.8           & 0.75           \\
                             & PSPEM & \textbf{78.9}          & \textbf{0.89} & \textbf{57.8}  & \textbf{0.19}  \\
\bottomrule
\end{tabular}
\end{table}

\section{CONCLUSION}
This paper presents PSPEM, an innovative prompt-based knowledge editing method, which seamlessly integrates a two-step process of compression and refinement to accurately extract and utilize crucial information from prompts. By employing alignment techniques, PSPEM ensures the generated text remains in harmony with the intended prompts, combining the high editing success of the weight-modified method with the ability to reason from given knowledge. Our experiments demonstrate PSPEM's robust performance in knowledge editing and attribute inserting tasks, notably highlighting its extraordinary advantage in imitating the influence of original prompts on the internal of the model.

The findings underscore PSPEM's potential to significantly advance the application of prompt engineering in knowledge editing, setting the stage for the evolution of more intuitive and precise language model (LM) editing tools. By facilitating a deeper alignment between model outputs and human-intended meanings, PSPEM not only enhances the accuracy of knowledge representation within LMs but also broadens the scope for their application across diverse domains. We anticipate that our contributions will act as a catalyst for further research in this area, ultimately leading to the development of more user-friendly and accurate LM editing tools.

\bibliographystyle{splncs04}
\bibliography{ref1}

\end{document}